\documentclass[a4paper]{article}

\usepackage[english]{babel}
\usepackage[utf8x]{inputenc}
\usepackage[T1]{fontenc}

\usepackage[a4paper,top=3cm,bottom=2cm,left=3cm,right=3cm,marginparwidth=1.75cm]{geometry}

\usepackage{multicol}
\usepackage{amsmath}
\usepackage{amssymb}
\usepackage{graphicx}
\usepackage{authblk}
\usepackage{algorithm2e}
\SetKwComment{Comment}{/* }{ */}
\RestyleAlgo{ruled}
\usepackage[colorinlistoftodos]{todonotes}
\usepackage[colorlinks=true, allcolors=blue]{hyperref}

\usepackage{booktabs}

\DeclareMathOperator*{\argmax}{argmax}
\title{Detection Selection Algorithm: A Likelihood based Optimization Method to Perform Post Processing for Object Detection}
\author{Angzhi Fan, Benjamin Ticknor and Yali Amit}
\affil{Department of Statistics, University of Chicago}
\begin{document}
\maketitle

\begin{abstract}
In object detection, post-processing methods like Non-maximum Suppression (NMS) are widely used. NMS can substantially reduce the number of false positive detections but may still keep some detections with low objectness scores. In order to find the exact number of objects and their labels in the image, we propose a post processing method called Detection Selection Algorithm (DSA) which is used after NMS or related methods. DSA greedily selects a subset of detected bounding boxes, together with full object reconstructions that give the interpretation of the whole image with highest likelihood, taking into account object occlusions. The algorithm consists of four components. First, we add an occlusion branch to Faster R-CNN to obtain occlusion relationships between objects. Second, we develop a single reconstruction algorithm which can reconstruct the whole appearance of an object given its visible part, based on the optimization of latent variables of a trained generative network which we call the decoder. Third, we propose a whole reconstruction algorithm which generates the joint reconstruction of all objects in a hypothesized interpretation, taking into account occlusion ordering. Finally we propose a greedy algorithm that incrementally adds or removes detections from a list to maximize the likelihood of the corresponding interpretation. DSA with NMS or Soft-NMS can achieve better results than NMS or Soft-NMS themselves, as is illustrated in our experiments on synthetic images with mutiple 3d objects.
\end{abstract}


\section{Introduction}
\label{sec:intro}

Object detection is one of the most important tasks in computer vision. Most object detection algorithms return detections in the form of bounding boxes. They give the center, height and width of each predicted bounding box, and perform classification of the object inside it. Some object detection algorithms, for example Faster R-CNN \cite{faster}, also provide an objectness score, which quantifies the confidence of having an object in the bounding box.

Usually, object detection algorithms first generate excessive detections and then use post processing methods like Non-maximum Suppression (NMS) to reduce the number of detections. NMS keeps the most promising detections through local comparisons. A NMS-threshold $N_t \in [0,1]$ is needed to determine when to suppress the less promising neighboring bounding boxes. After NMS, the remaining bounding boxes do not necessarily have high objectness scores and there are usually more bounding boxes than the real number of objects. In Faster R-CNN \cite{faster}, the top-N bounding boxes after NMS are declared as the final detections. But people usually don't know how many objects are in the image, so $N$ is still usually larger than the actual number of objects. 

Unlike NMS, Soft-NMS \cite{softNMS} does not directly eliminate less promising neighboring bounding boxes. It reduces their objectness scores as a function of the Intersection-over-Union (IoU). After Soft-NMS, we still need to use either a maximal number of boxes, or a lower bound threshold on the objectness socres of detections. The Soft-NMS \cite{softNMS} paper uses top $400$ detections per image on MS-COCO. 

To find the correct number of objects and labels in the image, one natural idea is to use a threshold $T$ on the objectness scores. Bounding boxes above $T$ are used as the final detections. The threshold $T$ can be determined by a validation set. But what if the validation set has a different distribution than the test set? The result can be very sensitive to the threshold $T$. A distribution shift from the validation set to the test set may cause significant damage to model performance. 

Our work proposes a novel post processing method for object detection algorithms building on the work in \cite{amit2007pop}. The idea is to find the most likely {\it interpretation} of an image, where an interpretation is an ordered subset of detections, ordered according to occlusion. Each object class is modeled by a generative model that maps a low dimensional latent space to image space, and the pixel values are assumed to be independent Gaussian conditional on the latent variables. The generative model provides both a reconstruction of the object image and a region of object support. The log-likelihood of the entire image conditional on the objects, their locations and the values of the latent variables is the sum of the log-likelihoods of the individual objects {\it on their visible parts}. That is why the occlusion ordering is essential; objects in the back are only visible outside the support of objects in front. Optimizing over ordered subset of detections is prohibitive computationally. We thus use a greedy search, where we take advantage of the objectness scores provided by the Faster R-CNN. Furthermore we add an additional branch to the Faster R-CNN that provides an occlusion score between 0 and 1, we call this Faster R-CNN-OC. And object with higher occlusion score is assumed in front of one with lower occlusion score. These outputs provide important inputs to the greedy search for the most likely interpretation as described below. In short,
the Detections Selection Algorithm (DSA) is a greedy search over ordered subsets of detections for the one with highest likelihood, using objectness and occlusion scores provided by the Faster R-CNN.

The first component of DSA is a single reconstruction algorithm that reconstructs the entire object based on its visible parts. Here we use a decoder architecture as in VAE's, and instead of reconstructing based on latent variables predicted from an encoder, we optimize over the latent variables. This modification is essential because we don't always see the whole object, and the reconstruction loss - the negative log-likelihood (NLL) is computed only on the visible part. It is not possible to adapt the encoder to different visible inputs. The single reconstruction algorithm takes a bounding box and its associated information, as well as reconstructions of previous objects in the sequence as input, and returns the whole appearance of the hypothesized object for that bounding box. The output is an approximation of the negative marginal log-likelihood of the image data in the visible part. A byproduct of the single reconstruction algorithm is that it performs amodal instance segmentation using the reconstruction. The second component of DSA is the whole reconstruction algorithm. It puts together the results of the single reconstruction algorithm on the current sequence of selected objects to provide the NLL of the entire image data, given this sequence. The final component - greedy search, is a greedy search through the detections provided by the Faster R-CNN-OC, ordered based on their objectness score. At each step we add one detection, reconstruct it from its visible part - computed as the complement within its bounding box of the union of supports of previously reconstructed detections with {\it higher occlusion scores}. If the NLL of the whole reconstruction is lower we include the additional object, if not we omit it. We also perform a one step- back search, in case eliminating a previously added object decreases the NLL if the new object is included. We add a fixed penalty term for each added object which is equivalent to an exponential prior on the number of objects. At some point the decrease in NLL of an additional object is cancelled by the object penalty and the search terminates. 

The main contribution of this paper is the DSA algorithm consisting of the three components described above. Our main goal is to find the exact number of objects and labels in the image, based on the whole reconstruction with the lowest NLL. There are several byproducts:
\begin{itemize}
    \item The augmented Faster R-CNN-OC provides an occlusion ordering. In particular we have found that it is sufficient to train the Faster R-CNN-OC only on {\it pairs} of objects to get excellent detections on multiple object scenes together with very reliable occlusion scores.
    
    \item The single reconstruction algorithm automatically reconstructs the invisible parts of objects and can be easily used for amodal instance segmentation.
    
    \item The whole reconstruction algorithm provides a way to generate an image given several hypothesized objects and their locations.
\end{itemize}

The rest of this article is organized as follows: section \ref{sec:relatedwork} introduces some related work about object detection, post processing methods, occlusion relationship reasoning and amodal instance segmentation. The probabilistic framework and three algorithms are explained in section \ref{sec:method}. Our dataset is described in section \ref{sec:data} and experiments are shown in section \ref{sec:exp}. Finally we provide a  discussion in section \ref{sec:conc}. 

\section{Related Work}
\label{sec:relatedwork}

The application of neural networks in object detection has attracted much attention. Some object detection algorithms rely on region proposal generation. For example, Fast R-CNN \cite{FastRCNN} and Faster R-CNN \cite{faster}. Some are single-stage methods, including Single Shot MultiBox Detector (SSD) \cite{SSD} and You Only Look Once (YOLO) \cite{YOLO}. All these object detection methods predict class probabilities and bounding box locations. 
%A bounding box's location can be characterized by four numbers: two numbers for its center, one number for its height and one number for its width. 
Faster R-CNN \cite{faster} also predicts the objectness score which represent the confidence of a detection. 

Post-processing is an important step to remove false positive detections in all object detection algorithms. One type of popular post-processing method is Non-maximum Suppression (NMS) and its variants. The paper Efficient Non-Maximum Suppression \cite{ENMS} proposed several algorithms to accelerate NMS. A recent review paper \cite{review_NMS} summarized five NMS techniques: Soft-NMS \cite{softNMS}, Softer-NMS \cite{softerNMS}, IOU-Guided NMS \cite{IoU-GuidedNMS}, Adaptive NMS \cite{AdaptiveNMS} and DIoU-NMS \cite{DIOU-NMS}. 

These five methods emphasize local information as opposed to optimizing a global objective function. Among them, Softer-NMS \cite{softerNMS}, IOU-Guided NMS \cite{IoU-GuidedNMS} and Adaptive NMS \cite{AdaptiveNMS} require modifying the detection model or adding additional modules. Instead of setting  a threshold to suppress highly overlapping bounding boxes, in each step Soft-NMS \cite{softNMS} decreases the detection score by a factor that depends on the IoU. Distance-IoU (DIoU) \cite{DIOU-NMS} takes into account the distance between the centers of bounding boxes. The idea of DIoU can be used in  NMS and in designing IoU-related loss functions.

Another type of post-processing method defines a global objective functions and uses some search procedure to choose the final detections. Examples are a Bayesian model for face detection \cite{NMS-face}, HS-NMS \cite{HS-NMS} and  probabilistic faster R-CNN \cite{probabilisticFRCNN}. Probabilistic faster R-CNN  \cite{probabilisticFRCNN} trains Gaussian Mixture Models (GMM) on heights and widths of region proposals \cite{FastRCNN} and uses GMM to calculate the likelihood for each region proposal.

The Bayesian model in \cite{NMS-face} first uses a kernel smoother on the face hypotheses to estimate the prior distribution, then uses face templates to estimate face likelihood, and use MCMC to get a stable face distribution from the posterior distribution. Our work differs in that we try to find the correct number of objects and take into account the occlusion relationships of objects.

Understanding the occlusion relationship between objects is called Occlusion Relationship Reasoning. MT-ORL \cite{MT-ORL} can predict object boundary maps and occlusion orientation maps and requires corresponding ground truths in order to train. A recent work \cite{ORM} performs occlusion relationship reasoning by pixel-level competition for conflict areas in segmentation. 

Researchers have payed attention to reconstructing the invisible parts and predicting the entire mask of an object including its invisible parts. The latter is often called amodal instance segmentation. SeGAN \cite{seGAN} jointly predict invisible masks and generate invisible parts of objects under the GAN \cite{GAN} framework. One work \cite{KINS} uses Multi-Level Coding to guide invisible mask prediction by multi-branch features. 

Our idea to compare likelihoods of the image based on different hypothesis of objects is motivated by POP model \cite{amit2007pop}. However the POP model uses a deformable template to model objects, as opposed to the more flexible decoder structure, and needs to find the occlusion ordering of the objects as part of the optimization. Here we take advantage of the output of the Faster R-CNN-OC to obtain the occlusion ordering. Furthermore we apply the algorithm to images of 3d objects, whereas the POP model was restricted to 2d objects. 
%Similar to Deformable Classifiers \cite{shen2019deformable}, we use Spatial Transformer Network %\cite{SPN} to transform images.

\section{Our Method}
\label{sec:method}

For simplicity, we assume the image $I$ consists of some objects placed on a blank background. We run a detection algorithm like the Faster R-CNN-OC as detailed in Section \ref{sec:ob} to get detections. The detections are in the form of $\{\textbf{det}_{i}=(score_{i}, bb_{i}, occ_{i}, cls_{i})\}_{i=1}^N$.  For each detection $\textbf{det}_{i}$, $score_{i}$ is the objectness score defined in \cite{faster}, $bb_{i}$ is the bounding box, $cls_{i}$ is the classification result, and $occ_{i}$ is the occlusion score obtained from the occlusion branch. We denote by $I[bb_{i}]$ the image restricted to the bounding box $bb_{i}$.
If the objects in $\textbf{det}_{i}$ and  $\textbf{det}_{j}$ overlap, and $occ_i<occ_j$, the object in $\textbf{det}_{i}$ is predicted to be occluded in the overlapping area by the object in $\textbf{det}_{j}$. 
%However, by many object detection algorithms like the original Faster R-CNN, each detection only includes $(score_{i}, bb_{i}, cls_{i})$. 
Typically there are false positive detections among all detections $\{(score_{i}, bb_{i}, cls_{i})\}_{i=1}^{N}$ produced by the Faster R-CNN-OC. In other words, only a subset of $\{(score_{i}, bb_{i}, cls_{i})\}_{i=1}^{N}$ is correct. Our goal is to find the ordered subset $\{\textbf{det}_{i_j}=(score_{i_j}, bb_{i_j}, occ_{i_j}, cls_{i_j})\}_{j=1}^k$ which yields the best interpretation of the image, namely the lowest NLL. But, trying every possible ordered subset of $\{\textbf{det}_{i}\}_{i=1}^N$ is  computational prohibitive. Our proposed Detections Selection Algorithm (DSA) in Section \ref{sec:dsa} greedily selects the detections when processing them according to their objectness scores from high to low, taking into account the occlusion scores to identify the visible parts of each object.

\subsection{Faster R-CNN-OC}
\label{sec:ob}

Faster R-CNN \cite{faster} is a detection framework which uses a Region Proposal Network (RPN) to generate region proposals, and uses the Fast-RCNN \cite{girshick2015fast} module to do bounding box regression and  classification for each region proposal. From the RPN we get an objectness score for each bounding box, with a higher score indicating more confident detection.

\begin{figure}[t]
  \centering
  \includegraphics[width=\linewidth]{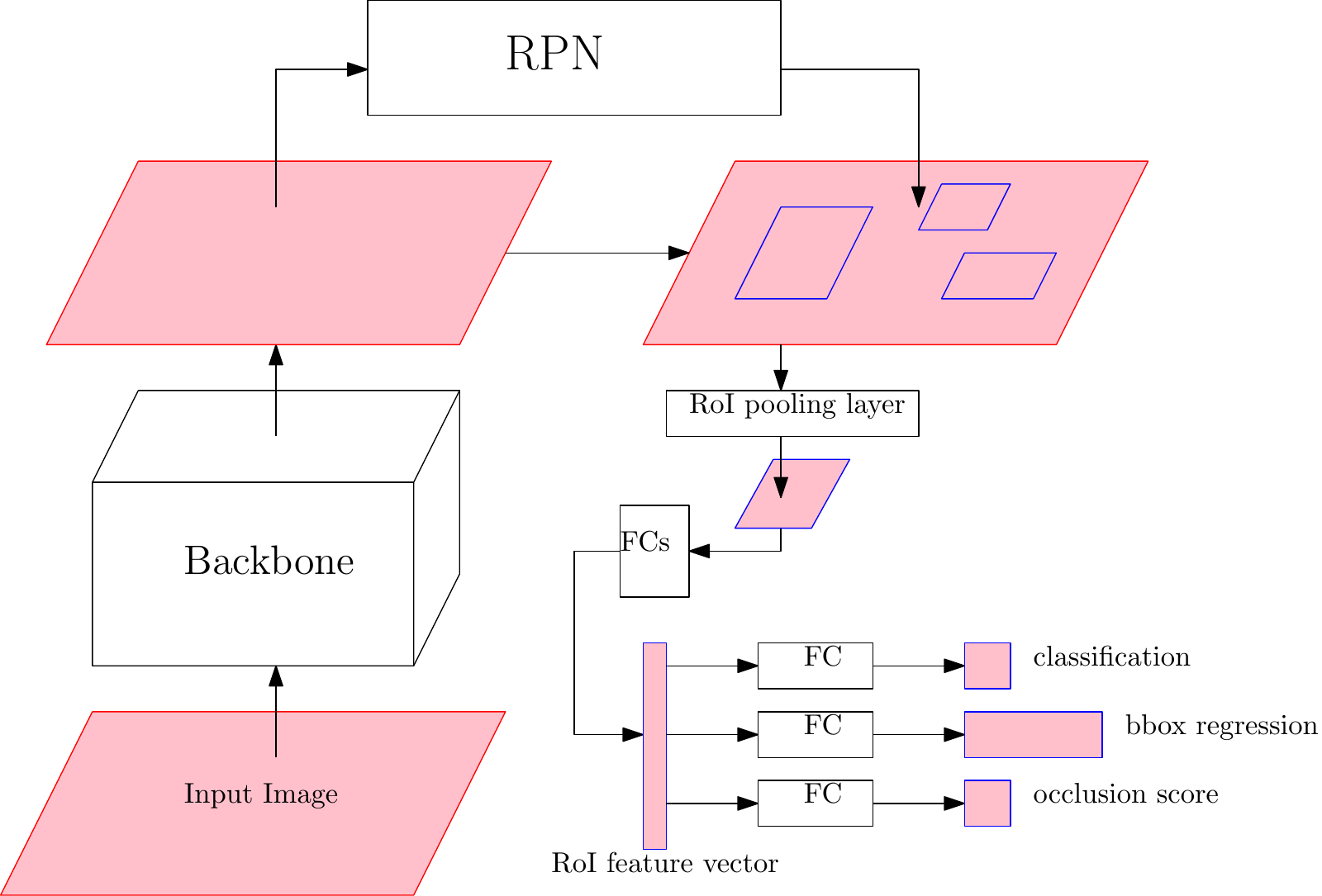}
  \caption{Faster R-CNN with occlusion branch.}
  \label{fig:ob}
\end{figure}

We added an additional branch called the occlusion branch to Faster R-CNN parallel to the regression branch and the classification branch inside the Fast-RCNN module. The  Faster R-CNN-OC is shown in Figure \ref{fig:ob}. 

The occlusion branch also consists of a fully-connected layer. The output of the fully-connected layer goes through a sigmoid function so that we get an occlusion score between 0 and 1. As we mentioned earlier, if two objects overlap, the one with higher occlusion score is predicted to be visible in the overlapping area. During inference, we only need to compare the occlusion scores of objects to determine the occlusion sequence. One way to train the occlusion branch is to provide pairs of overlapping objects, and assign the occlusion score of the occluded and occluding object to be 0 and 1 respectively. More details will be explained in Section \ref{exp:imple}.

This occlusion branch has provided reliable results in our experiments. We trained it on two-object images, and found that it generalizes well to images which contain multiple objects. Related experiments can be found in Section \ref{exp:occ}.

\subsection{Single Reconstruction Algorithm}
\label{sec:sr}
Our single reconstruction algorithm is based on a generative model of the form of a decoder in a VAE \cite{kingma2013auto}.  During training of a full VAE we maximize the (variational) lower bound on the marginal likelihood

\begin{equation} 
\label{equ:elbo}
\mathcal{L}(\theta, \phi; x) = \mathbf{E}_{q_{\phi}(z|x)}(\log p_{\theta}(x|z))-\mathbf{D}_{KL}(q_{\phi}(z|x)||p(z))
\end{equation}
where $\phi$ and $\theta$ are encoder and decoder parameters. For simplicity, assume $z \sim \mathcal{N}(\mathbf{0},\mathbf{I}_{N_z})$, $x|z \sim \mathcal{N}(m_{\theta,z},\sigma^2 \mathbf{I})$ and $q_{\phi}(z|x)$ is the density of $\mathcal{N}(\mu_{x},\Gamma_{x})$, where $m_{\theta,z}$ is the output of the decoder if the input is $z$, and $\Gamma_{x}$ is a diagonal matrix
\begin{equation} 
  \Gamma_{x} =
  \begin{bmatrix}
    \tau_{x,1}^2 & & \\
    & \ddots & \\
    & & \tau_{x,N_z}^2
  \end{bmatrix}
\end{equation}
with $\tau_{x,i}>0$ for $i=1,2,...,N_z$.

During inference part of the object may be occluded, and the reconstruction will be based only on the visible part which makes it difficult for the encoder to predict $\mu_{x}$ and $\Gamma_{x}$.  Instead, both in training and in inference, we optimize over $\mu_{x}$ and $\Gamma_{x}$ while fixing the decoder parameters $\theta$. Thus we drop the VAE encoder and only train  the decoder. 
After  training, we can find $\mu_{x}$ and $\Gamma_{x}$ for an incomplete object, then pass $z=\mu_{x}$ to the decoder to reconstruct the whole appearance of the object. 

The first half of this section assumes the target bounding box has the same size as the decoder training images. But in reality the target bounding boxes may have different sizes. So we use the parameterised sampling grid idea of Spatial Transformer Networks \cite{SPN} to deal with this issue as is explained in the second half of this section.  

\subsubsection{Fixed-size Reconstructions}
The decoder is trained using the method of stochastic variational inference (SVI) \cite{hoffman2013stochastic}. Instead of using the encoder to predict $\mu_x$, $\Gamma_x$, we update these variables with a fixed number of optimization steps using gradient descent. Then we fix $\mu_x$, $\Gamma_x$ and update the decoder parameters $\theta$. We do these updates iteratively until convergence. The training data of our decoder are fixed-size images. Each image only has one object in it and we train a separate model for each class.

During reconstruction, for a target bounding box $\Tilde{x}$, where the set of visible pixels is $V$, we optimize $\mu_{\Tilde{x}}$, $\Gamma_{\Tilde{x}}$ to maximize Equation \ref{equ:elbo} based on the visible part, while keeping the decoder fixed. To be more specific, Equation \ref{equ:elbo} becomes

\begin{equation} 
\begin{split}
\mathcal{L}(\theta, \phi; \Tilde{x}, V) = & \mathbf{E}_{q_{\phi}(z|\Tilde{x})}(\log p_{\theta,V}(\Tilde{x}|z))-\mathbf{D}_{KL}(q_{\phi}(z|\Tilde{x})||p(z))\\
= & \mathbf{E}_{\mathcal{N}(\mu_{\Tilde{x}},\Gamma_{\Tilde{x}})}(\log p_{\theta,V}(\Tilde{x}|z))-\mathbf{D}_{KL}(\mathcal{N}(\mu_{\Tilde{x}},\Gamma_{\Tilde{x}})||\mathcal{N}(\mathbf{0},\mathbf{I}_{N_z}))\\
= & \mathbf{E}_{\mathcal{N}(\mu_{\Tilde{x}},\Gamma_{\Tilde{x}})}(- \frac{|V|}{2}\log(2\pi \sigma^2)-\frac{1}{2\sigma^2}\sum_{i\in V}(m_{\theta,z,i}-\Tilde{x}_i)^2)\\
&-(\frac{\|\mu_{\Tilde{x}}\|_2^2+\displaystyle\sum_{i=1}^{N_z} \tau_{\Tilde{x},i}^2-N_z}{2}-\sum_{i=1}^{N_z}\log\tau_{\Tilde{x},i})
\end{split}
\end{equation}
then we can get our reconstruction as
\begin{equation} \hat{\Tilde{x}}=m_{\theta,z_{\Tilde{x}}^*}
\end{equation} 
where $z_{\Tilde{x}}^*$ is set to be $\mu_{\Tilde{x}}$. Since the decoder is trained with complete objects, the output will contain complete objects.

\subsubsection{Arbitrary-size Reconstructions}
\label{subsubsec:arb_size_recon}
\begin{figure}[t]
  \centering
  \includegraphics[width=\linewidth]{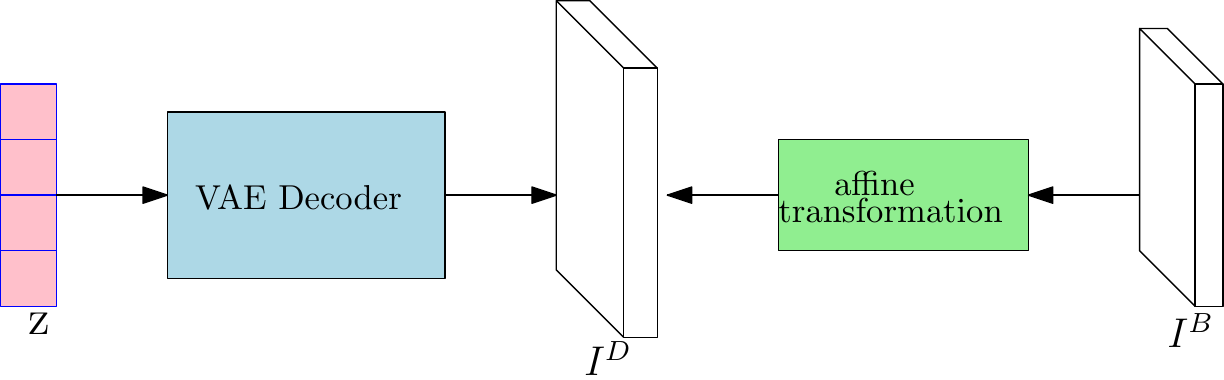}
  \caption{Decoder architecture.}
  \label{fig:decoder}
\end{figure}

Our target image is the visible part of everything inside the bounding box. But the bounding box size is usually not the same as the size of VAE training images. To deal with this issue, we turn to the parameterized sampling grid idea of the Spatial Transformer Network \cite{SPN}. A similar idea also appears in \cite{gregor2015draw}. 
%In the Spatial Transformer Network \cite{SPN}, an affine transformation 
% \begin{equation} 
% \begin{pmatrix}
% x_i^s \\
% y_i^s 
% \end{pmatrix} =\begin{pmatrix}
% \theta_{11} & \theta_{12} & \theta_{13}\\
% \theta_{21} & \theta_{22} & \theta_{23}
% \end{pmatrix} \begin{pmatrix}
% x_i^t \\
% y_i^t \\
% 1
% \end{pmatrix}
% \end{equation}
% sets up the correspondence between target coordinate $(x_i^t,y_i^t)$ in the output feature map and source coordinate $(x_i^s,y_i^s)$ in the input feature map. Ideally, coordinate $(x_i^t,y_i^t)$ in the output should have the same image value as coordinate $(x_i^s,y_i^s)$ in the input.
If our target image box is $I_{i_j}$, we assume an affine transformation between $I_{i_j}$ and the decoder output. For any coordinate $(x^b,y^b)$ in our reconstruction, after the affine transformation we get 

\begin{equation} 
\begin{pmatrix}
x^d \\
y^d 
\end{pmatrix} =\begin{pmatrix}
s_x & 0 & t_x s_x\\
0 & s_y & t_y s_y
\end{pmatrix} \begin{pmatrix}
x^b \\
y^b \\
1
\end{pmatrix}
\end{equation} 
and for each channel, the coordinate $(x^b,y^b)$ in our reconstruction should have the same image value as coordinate $(x^d,y^d)$ in the decoder output. As shown in Figure \ref{fig:decoder}, if our reconstruction is $I^{B}$ and the decoder output is $I^{D}$, and $I^{B}(m,n,c)$ or $I^{D}(m,n,c)$ is the pixel value at coordinate $(m,n)$ and channel $c$ in image $I^{B}$ or $I^{D}$, then 

\begin{equation} 
I^{D}(x_i^d, y_i^d,c) = I^{B}(x_i^b,y_i^b,c)
\end{equation} 

For simplicity, and to avoid too much flexibility, we do not consider rotation in the affine transformation. The shearing parameters $s_x, s_y$ represent scaling in the $x$ and $y$ axis. We assume  an isotropic scaling and fix $s_x=s_y=d/L$, where $(d,d,3)$ is the VAE training image size and $L$ is the maximum between the height $H_{I_{i_j}}$ and width $W_{I_{i_j}}$ of the target image box $I_{i_j}$. 
%The constant $d/L$ is chosen due to the fact that $x_i^b$ or $y_i^b$ may range from $1$ to $L$, but the %decoder output can only has size $(d,d,3)$. 
The $t_x, t_y$ translation parameters are kept free. 
%They are multiplied with $s_x, s_y$ because we want to disentangle $t_x, t_y$ from scaling factors.
The coordinate $(x^b,y^b)$ can be any integer coordinate in our reconstruction. However, after the transformation, we get the corresponding $(x^d,y^d)$, which may not be integers. We utilize the bilinear sampling kernel to interpolate for coordinate $(x^d,y^d)$. The bilinear sampling kernel is formulated as

% \begin{equation} 
% V_i^c = \sum_{n}^{H}\sum_{m}^{W} U_{nm}^c \max(0, 1-|x_i^s-m|)\max(0, 1-|y_i^s-n|)
% \end{equation} 
% where $V_i^c$ is the value of channel $c$ at coordinate $(x_i^t,y_i^t)$ in the output feature map, and $U_{nm}^c$ is the value of channel $c$ at coordinate $(m,n)$ in the input feature map.

% By our notations, the bilinear sampling kernel is
\begin{equation} 
I^B(x,y,c) = \sum_{m=1}^{d}\sum_{n=1}^{d} I^D(m,n,c) \max(0, 1-|x-m|)\max(0, 1-|y-n|)
\end{equation} 
 In this case, our reconstruction is an $L \times L \times 3$ image, so we need to crop a $H_{I_{i_j}}\times W_{I_{i_j}}\times 3$ region at the center of the $L \times L \times 3$ image to get our reconstruction for the target bounding box. 

With the affine transformation and bilinear sampling kernel, gradients can still back-propagate from the target image to the latent code. In this way, we compare the difference between the target image and our reconstruction for the target bounding box, and obtain the latent codes as well as $t_x$, $t_y$ using gradient descent.

In summary, for target image box $I_{i_j}$, we fix $s_x, s_y$ but optimize $\mu_{i_j}$, $\tau_{i_j}$ and $(t_x, t_y)$, where $\mu_{i_j}$ and $\tau_{i_j}$ are parameters in the posterior distribution $z|I_{i_j}, V_{i_j} \sim \mathcal{N}(\mu_{i_j},\Gamma_{i_j})$. Given $\mu_{i_j}$, $\Gamma_{i_j}$, $(t_x, t_y)$ and $(s_x, s_y)$, we sample $z$ from $\mathcal{N}(\mu_{i_j},\Gamma_{i_j})$, pass it to the decoder to get a $d\times d \times 3$ decoder output, and use the affine transformation and bilinear sampling kernel to get a $L \times L \times 3$ reconstruction called $R_{i_j}$. This procedure from $z$, $(t_x, t_y)$ and $(s_x, s_y)$ to $R_{i_j}$ is called Decoder($z$, $(t_x, t_y)$, $(s_x, s_y)$) in Algorithm \ref{alg:sr}.

\begin{algorithm}
\caption{Single Reconstruction Algorithm}\label{alg:sr}
\KwIn{Cropped target image $I_{i_j}=I[bb_{i_j}]$, $V_{i_j}$: the coordinates of the visible pixels in $I_{i_j}$, the detection $\textbf{det}_{i_j}=(score_{i_j}, bb_{i_j}, occ_{i_j}, cls_{i_j})$, VAE training image size $(d,d,3)$, $N_{iter}$, $N_{z}$ and $\sigma$ }
$L  \gets max(H_{I_{i_j}}, W_{I_{i_j}})$ \;
$\mu_{i_j} \gets zeros(N_{z})$
\Comment*[r]{$N_{z}$ is the dimension of the latent code}
$(\log \tau_{i_j,1}, \log \tau_{i_j,2},..., \log \tau_{i_j,N_{z}} ) \gets zeros(N_{z})$ \Comment*[r]{$(\tau_{i_j,1}^2, \tau_{i_j,2}^2,..., \tau_{i_j,N_{z}}^2 )$ is the diagonal of covariance matrix $\Gamma_{i_j}$}
$(t_x, t_y) \gets (0,0)$ \Comment*[r]{translation parameters}
$(s_x, s_y) \gets (d/L, d/L)$ \Comment*[r]{shearing parameters, fixed}

\For{$j=1$ \KwTo $N_{iter}$}{
  $z \gets$ sampled from $ \mathcal{N}(\mu_{i_j},\Gamma_{i_j})$ \;
  $R_{i_j} \gets$ Decoder($z$, $(t_x, t_y)$, $(s_x, s_y)$)
  \Comment*[r]{$R_{i_j}$ has size $(L,L,3)$}
  $R_{i_j}^{(bb)}  \gets$ Cropped region of size $(H_{I_{i_j}}, W_{I_{i_j}},3)$ at the center of $R_{i_j}$ \;
  $Loss \gets \mathbb{D}_{KL}(\mathcal{N}(\mu_{i_j}, \Gamma_{i_j})||\mathcal{N}(0, \mathbf{I}_{N_z}))+\frac{1}{2\sigma^2}\sum_{x \in V_{i_j}}(R_{i_j, x}^{(bb)}-I_{i_j, x})^2$ \;
  Update $\mu_{i_j}$, $(\log \tau_{i_j,1}, \log \tau_{i_j,2},..., \log \tau_{i_j,N_{z}} )$ and $(t_x, t_y)$ based on gradients of $Loss$ \;
    }
$R_{i_j} \gets$ Decoder($\mu_{i_j}$, $(t_x, t_y)$, $(s_x, s_y)$) \;
$bb_{i_j}^* \gets $ an $L \times L \times 3$ bounding box which centers at the center of bounding box $bb_{i_j}$\;
\KwOut{Single reconstruction $R_{i_j}$, parameters $\mu_{i_j}, \Gamma_{i_j}$ and the inferred bounding box $bb_{i_j}^*$}
\end{algorithm}
Besides $R_{i_j}$, Algorithm \ref{alg:sr} also returns $bb_{i_j}^*$, an $L \times L \times 3$ bounding box which centers at the center of the target bounding box $bb_{i_j}$. Because $R_{i_j}$ has size $L \times L \times 3$, if we place $R_{i_j}$ on the whole image, we should place it inside the bounding box $bb_{i_j}^*$, not $bb_{i_j}$. 

Our single reconstruction $R_{i_j}$ is an object on a blank background, and the support of the object is assumed to be all the pixels whose magnitude is larger than some pre-determined threshold $t_0$ called the "occlusion threshold". Since Algorithm \ref{alg:sr} aims at reconstructing the single object based on a single bounding box, we call it the single reconstruction algorithm.

\begin{figure}[t]
  \centering
  \includegraphics[width=\linewidth]{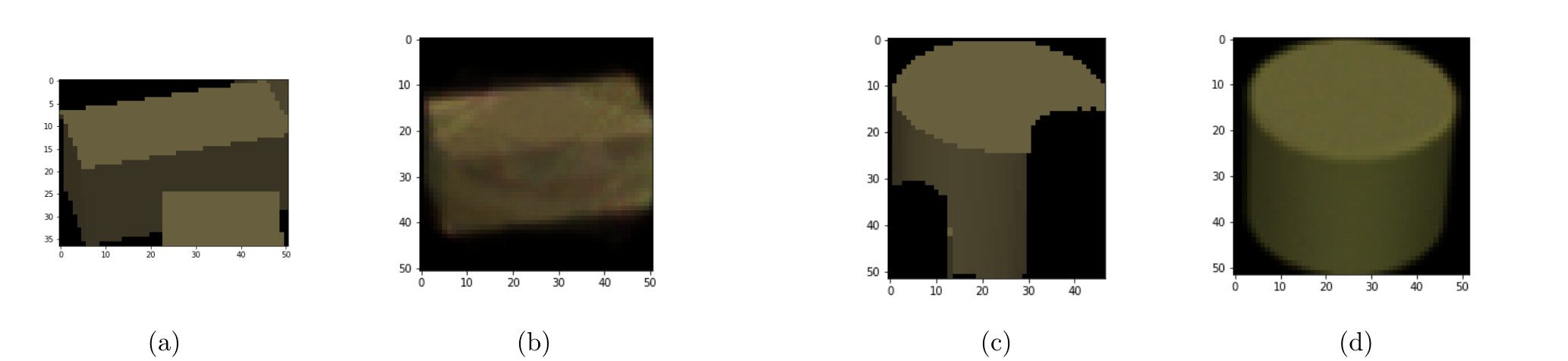}
  \caption{Example of the Single Reconstruction Algorithm. (a), (c) - target images with occluded parts in black, (b), (d) - reconstructions $R_{ij}$ on the $L \times L$ grid.}
  \label{fig:sr1}
\end{figure}

An example of the single reconstruction algorithm can be found in Figure \ref{fig:sr1}, where $d=50$. Image $(a)$ is the cropped target image where clutter is present. Image $(b)$ is the $L \times L \times 3$ single reconstruction for target $(a)$, where $L>50$. Similarly, image $(d)$ is the single reconstruction for target $(c)$, except that $(c)$ is now an incomplete image. 

\subsection{Whole Reconstruction Algorithm}
\label{sec:wr}
The whole reconstruction algorithm, see
Algorithm \ref{alg:wr}, is used to put together single reconstructions of  an ordered subset of detections  on a blank background the same size as the image,  according to their occlusion scores. As we mentioned earlier, if the supports of two objects overlap, the visible object in the overlapping area is the one with higher occlusion score.

In Algorithm \ref{alg:wr}, we loop through the ordered set of detections, if the single reconstruction of the current detection has not been computed we implement the single reconstruction algorithm. This requires computing which pixels within the current bounding box are visible, so we consider all pre-computed single reconstructions which have a higher occlusion score. We put those objects with higher occlusion scores on the background. After that, the blank part on the background is assumed to be still visible. 
\begin{algorithm}
\caption{Whole Reconstruction Algorithm}\label{alg:wr}
\KwIn{
A subset of detection results $\{\textbf{det}_{i_j}=(score_{i_j}, bb_{i_j}, occ_{i_j}, cls_{i_j})\}_{j=1}^k$, reconstructions hashmap $ReconDict$, occlusion threshold $t_0$}
Sort $\{\textbf{det}_{i_j}\}_{j=1}^k$ according to $occ_{i_j}$ from high to low\;
$Canvas \gets zeros(H,W,3)$
\Comment*[r]{$(H,W,3)$ is the image size}
\For{$j=1$ \KwTo $k$}{
  \If{$ReconDict[i_j]$ doesn't exist}{$I_{i_j} \gets I[bb_{i_j}]$ \Comment*[r]{ $I$ is the image and $I[bb_{i_j}]$ is obtained by cropping the image at bounding box $bb_{i_j}$}
  $V_{i_j} \gets$ The coordinates of blank pixels in $Canvas[bb_{i_j}]$ \Comment*[r]{"blank pixel" is a pixel with value $(0,0,0)$}
  $(R_{i_j}, \mu_{i_j}, \Gamma_{i_j}, bb_{i_j}^*) \gets SingleReconstruction(I_{i_j}, V_{i_j},\textbf{det}_{i_j})$\;
  $ReconDict[i_j] \gets (R_{i_j}, \mu_{i_j}, \Gamma_{i_j}, bb_{i_j}^*)$\;
  }
  $(R_{i_j}, \mu_{i_j}, \Gamma_{i_j}, bb_{i_j}^*) \gets ReconDict[i_j]$\;
  $l_{i_j}, r_{i_j}$ - coordinates of upper left hand corner of $bb_{i_j}^*$\;
  \For{pixel $x,y$ in $R_{i_j}$}{
    \If{$Canvas[l_{i_j}+x,r_{i_j}+y]$ is $(0,0,0)$ and the magnitude of $R_{i_j}[x,y]$ is above $t_0$ }{$Canvas[l_{i_j}+x,r_{i_j}+y]\gets R_{i_j}[x,y]$\;
    }}}
\KwOut{Whole reconstruction $Canvas$, reconstructions hashmap $ReconDict$}
\end{algorithm}

As mentioned in the single reconstruction algorithm, a pre-determined threshold $t_0$ is used to determine which pixels constitute the support of the object in its single reconstruction. Only those pixels which are within the support acquire the values of the reconstruction, the others remain blank.

\begin{figure}[t]
  \centering
  \includegraphics[width=\linewidth]{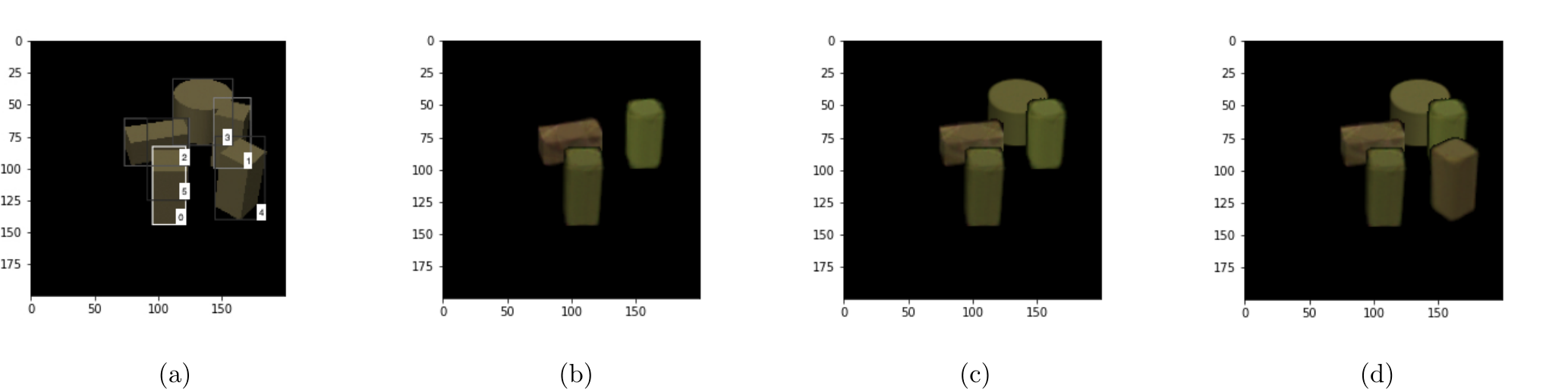}
  \caption{Example of Whole Reconstruction Algorithm.}
  \label{fig:wr1}
\end{figure}

Figure \ref{fig:wr1} provides an example of Algorithm \ref{alg:wr}. The original image with $5$ objects are shown in $(a)$. In $(a)$ there are $6$ detected bounding boxes labeled at their lower right corners from $0$ to $5$ according to their objectness scores from high to low. Image $(b), (c),  (d)$ are the whole reconstruction canvases when we select detections $[0,1,2]$, $[0,1,2,3]$ and $[0,1,2,3,4]$ respectively. Single reconstructions from Figure \ref{fig:sr1} are used here. From $(b)$ to $(c)$ the cylinder in bounding box $3$ is occluded by two cuboids in bounding boxes $1,2$. Image $(c)$ in Figure \ref{fig:sr1} is an incomplete cylinder because we only keep the visible part. 

It is worth noting that the single reconstruction has size $(L,L,3)$ and is usually larger than the size of the target bounding box. However, we don't restrict the single reconstruction inside the region of the target bounding box. In other words, if the support of the object is outside the target bounding box, we  still add those pixels to the canvas in the whole reconstruction algorithm. Our reason for doing this is that if the reconstructed object indeed exists, its part outside its target bounding box should exist as well.

\subsection{Probabilistic Framework}
\label{sec:prob}
In our approach a good interpretation of the image should achieve two goals. First, we want to get a good reconstruction of the whole image based on our selected detections. Second, we need to avoid selecting redundant detections. These two goals motivate the following probabilistic framework.

Suppose a detection algorithm yields detection results $\{\textbf{det}_{i}=(score_{i}, bb_{i}, occ_{i}, cls_{i})\}_{i=1}^N$, and $\{\textbf{det}_{i_j}=(score_{i_j}, bb_{i_j}, occ_{i_j}, cls_{i_j})\}_{j=1}^k$ is a subset used to interpret the image. For the number of detections $k$ in the subset, we assume a prior $p_{K}(k) \propto e^{-\lambda_0 k}$ for $k \geq 0$. For every detection $\textbf{det}_{i}$, its latent code $z_i$ has the same prior $z_i \sim \mathcal{N}(0, \mathbf{I}_{N_z})$. We assume non-informative prior on each $(score_{i}, bb_{i}, occ_{i}, cls_{i})$, so the prior for $\{\textbf{det}_{i_j}\}_{j=1}^k$ is equivalent to the prior $p_{K}(k)$, i.e. $p(\{\textbf{det}_{i_j}\}_{j=1}^k)=p_{K}(k)$.

Given $\{\textbf{det}_{i_j}=(score_{i_j}, bb_{i_j}, occ_{i_j}, cls_{i_j})\}_{j=1}^k$ and $\{z_{i_j}\}_{j=1}^k$, we assume the distribution of the hypothesized image is Gaussian with the same variance $\sigma^2$ on each pixel, and the pixels are independent conditioned on the mean. The mean of this Gaussian distribution is assumed to be the  reconstruction produced by the whole reconstruction algorithm. 

Therefore, if the image is $I$, the log marginal likelihood of the interpretation is 
\begin{equation} 
\begin{split}
\log p(I, \{\textbf{det}_{i_j}\}_{j=1}^k)&=\log p_{K}(k) +\log p(I|\{\textbf{det}_{i_j}\}_{j=1}^k)
\\&=\log p_{K}(k) + \log \int \int \cdots \int p(I|\{z_{i_j}\}_{j=1}^k, \{\textbf{det}_{i_j}\}_{j=1}^k) \prod_{j=1}^k p(z_{i_j}) d z_{i_1} d z_{i_2} \cdots d z_{i_k}
\end{split}
\end{equation} 
which is intractable in terms of computation. Using the well known variational approximation
\begin{equation} 
\begin{split}
& \log p(I|\{\textbf{det}_{i_j}\}_{j=1}^k)\\
= & \mathbf{E}_{q_{\phi}(\{z_{i_j}\}_{j=1}^k|I,\{\textbf{det}_{i_j}\}_{j=1}^k)}\log \frac{p(I,\{z_{i_j}\}_{j=1}^k|\{\textbf{det}_{i_j}\}_{j=1}^k)}{q_{\phi}(\{z_{i_j}\}_{j=1}^k|I,\{\textbf{det}_{i_j}\}_{j=1}^k)}\\&+\mathbf{D}_{KL}(q_{\phi}(\{z_{i_j}\}_{j=1}^k|I,\{\textbf{det}_{i_j}\}_{j=1}^k)||p(\{z_{i_j}\}_{j=1}^k|I,\{\textbf{det}_{i_j}\}_{j=1}^k))\\
= & \mathbf{E}_{q_{\phi}(\{z_{i_j}\}_{j=1}^k|I,\{\textbf{det}_{i_j}\}_{j=1}^k)}\log p(I|\{z_{i_j}\}_{j=1}^k, \{\textbf{det}_{i_j}\}_{j=1}^k)\\&-\mathbf{D}_{KL}(q_{\phi}(\{z_{i_j}\}_{j=1}^k|I,\{\textbf{det}_{i_j}\}_{j=1}^k)||p(\{z_{i_j}\}_{j=1}^k|\{\textbf{det}_{i_j}\}_{j=1}^k))\\
&+\mathbf{D}_{KL}(q_{\phi}(\{z_{i_j}\}_{j=1}^k|I,\{\textbf{det}_{i_j}\}_{j=1}^k)||p(\{z_{i_j}\}_{j=1}^k|I,\{\textbf{det}_{i_j}\}_{j=1}^k))
\end{split}
\end{equation} 
and 
\begin{equation} p(\{z_{i_j}\}_{j=1}^k|\{\textbf{det}_{i_j}\}_{j=1}^k)=p(\{z_{i_j}\}_{j=1}^k)
\end{equation} 
we drop the last Kullback–Leibler divergence term and use
\begin{equation} 
\label{equ:approx1}
\mathbf{E}_{q_{\phi}(\{z_{i_j}\}_{j=1}^k|I,\{\textbf{det}_{i_j}\}_{j=1}^k)}\log p(I|\{z_{i_j}\}_{j=1}^k, \{\textbf{det}_{i_j}\}_{j=1}^k)-\mathbf{D}_{KL}(q_{\phi}(\{z_{i_j}\}_{j=1}^k|I,\{\textbf{det}_{i_j}\}_{j=1}^k)||p(\{z_{i_j}\}_{j=1}^k))
\end{equation} 
to approximate $\log p(I|\{\textbf{det}_{i_j}\}_{j=1}^k)$. 

As in previous sections, $I_{i_j}=I[bb_{i_j}]$ denotes the cropped image from $I$ at bounding box $bb_{i_j}$. Without loss of generality, we can assume $\{\textbf{det}_{i_j}\}_{j=1}^k$ is sorted by occlusion scores from high to low. Then we use the following approximation

\begin{equation} 
\label{equ:approx1.5}
q_{\phi}(\{z_{i_j}\}_{j=1}^k|I,\{\textbf{det}_{i_j}\}_{j=1}^k)\approx \prod_{j=1}^k q_{\phi}(z_{i_j}|I_{i_j}, V_{i_j}, \textbf{det}_{i_j})
\end{equation} 
where $V_{i_j}$ denotes the visible pixels in the bounding box of object $i_j$ taking into account the union of supports of reconstructions $i_r$, where $r=1,\ldots,j-1$ (see  Algorithm \ref{alg:wr}), and $q_{\phi}(z_{i_j}|I_{i_j}, V_{i_j}, \textbf{det}_{i_j})$ is the posterior distribution of $z_{i_j}$ given cropped image $I_{i_j}$, $V_{i_j}$ and detection $\textbf{det}_{i_j}$. Using $z_{i_j}|I_{i_j}, V_{i_j}, \textbf{det}_{i_j} \sim \mathcal{N}(\mu_{i_j},\Gamma_{i_j})$, equation \ref{equ:approx1} can be approximated by \begin{equation} 
\label{equ:approx2}
\begin{split}
&\mathbf{E}_{\prod_{j=1}^k q_{\phi}(z_{i_j}|I_{i_j}, V_{i_j}, \textbf{det}_{i_j})}\log p(I|\{z_{i_j}\}_{j=1}^k, \{\textbf{det}_{i_j}\}_{j=1}^k)-\sum_{j=1}^k\mathbf{D}_{KL}(q_{\phi}(z_{i_j}|I_{i_j}, V_{i_j}, \textbf{det}_{i_j})||p(z_{i_j}))\\
=&\mathbf{E}_{\prod_{j=1}^k q_{\phi}(z_{i_j}|I_{i_j}, V_{i_j}, \textbf{det}_{i_j})}\log p(I|\{z_{i_j}\}_{j=1}^k, \{\textbf{det}_{i_j}\}_{j=1}^k)-\sum_{j=1}^k\mathbf{D}_{KL}(\mathcal{N}(\mu_{i_j},\Gamma_{i_j})||\mathcal{N}(0, \mathbf{I}_{N_z}))\\
=&\mathbf{E}_{\prod_{j=1}^k q_{\phi}(z_{i_j}|I_{i_j}, V_{i_j}, \textbf{det}_{i_j})}\log p(I|\{z_{i_j}\}_{j=1}^k, \{\textbf{det}_{i_j}\}_{j=1}^k)-\sum_{j=1}^k(\frac{\|\mu_{i_j}\|_2^2+\displaystyle \sum_{t=1}^{N_z} \tau_{i_j, t}^2-N_z}{2}-\sum_{t=1}^{N_z}\log\tau_{i_j, t})
\end{split}
\end{equation} 

In this way, we approximate the log marginal likelihood as
\begin{equation} 
\label{equ:approx3}
\begin{split}
&\log p(I,\{\textbf{det}_{i_j}\}_{j=1}^k)\\
\approx & \log p_{K}(k) + \mathbf{E}_{\prod_{j=1}^k q_{\phi}(z_{i_j}|I_{i_j}, V_{i_j}, \textbf{det}_{i_j})}\log p(I|\{z_{i_j}\}_{j=1}^k, \{\textbf{det}_{i_j}\}_{j=1}^k)\\
&-\sum_{j=1}^k(\frac{\|\mu_{i_j}\|_2^2+\displaystyle \sum_{t=1}^{N_z} \tau_{i_j, t}^2-N_z}{2}-\sum_{t=1}^{N_z}\log\tau_{i_j, t})\\
\approx &  \log p_{K}(k) + \log p(I|\{z_{i_j}^*\}_{j=1}^k, \{\textbf{det}_{i_j}\}_{j=1}^k)-\sum_{j=1}^k\frac{\|\mu_{i_j}\|_2^2+\displaystyle \sum_{t=1}^{N_z} \tau_{i_j, t}^2-N_z}{2}+\sum_{j=1}^k\sum_{t=1}^{N_z}\log\tau_{i_j, t}
\end{split}
\end{equation} 
where $z_{i_j}^*$ is sampled from $\mathcal{N}(\mu_{i_j},\Gamma_{i_j})$. In the single reconstruction algorithm, minimizing the loss gives us $\mu_{i_j}$, $\Gamma_{i_j}$ and $(t_x, t_y)$, so we can sample from $\mathcal{N}(\mu_{i_j},\Gamma_{i_j})$. If the whole reconstruction using $\{z_{i_j}^*\}_{j=1}^k$ and $ \{\textbf{det}_{i_j}\}_{j=1}^k$ is $I_{\{i_1,...,i_k\}}$, then 
\begin{equation}\label{lik}
    \log p(I|\{z_{i_j}^*\}_{j=1}^k, \{\textbf{det}_{i_j}\}_{j=1}^k)=- \frac{|I|}{2}\log(2\pi \sigma^2)-\frac{1}{2 \sigma^2}\|I-I_{\{i_1,...,i_k\}}\|_{vec,2}^2
\end{equation}

It is important to emphasize that \eqref{lik} always provides a log-likelihood for the entire image. The $Canvas$ output of the whole reconstruction algorithm provides the union of the supports of the different objects in the image as the set of all non-zero pixels. The complement of this set is considered background and the hypothesized distribution at each background pixel based on the equations above is simply $N(0,\sigma^2)$ in each channel.

\subsection{Detection Selection Algorithm - Greedy Search}
\label{sec:dsa}

 Note that $p_{K}(k)=\frac{e^{-\lambda_0 k}}{e^{\lambda_0}/(e^{\lambda_0}-1)}$ for $k\geq 0$. 
Based on our probabilistic framework, if $\{i_1,...,i_k\}$ are the indices of the selected detections, which are used as an interpretation of the image, we have
\begin{equation} 
\label{equ:logp_formula}
\begin{split}
&\log p(I,\{\textbf{det}_{i_j}\}_{j=1}^k)\\
\approx & \log p_{K}(k) + \log p(I|\{z_{i_j}^*\}_{j=1}^k, \{\textbf{det}_{i_j}\}_{j=1}^k)-\sum_{j=1}^k\frac{\|\mu_{i_j}\|_2^2+\displaystyle \sum_{t=1}^{N_z} \tau_{i_j, t}^2-N_z}{2}+\sum_{j=1}^k\sum_{t=1}^{N_z}\log\tau_{i_j, t}\\
=& \log \frac{e^{-\lambda_0 k}}{e^{\lambda_0}/(e^{\lambda_0}-1)}-\frac{|I|}{2}\log(2\pi \sigma^2) -\frac{1}{2 \sigma^2}\|I-I_{\{i_1,...,i_k\}}\|_{vec,2}^2\\
&-\sum_{j=1}^k\frac{\|\mu_{i_j}\|_2^2+\displaystyle \sum_{t=1}^{N_z} \tau_{i_j, t}^2-N_z}{2}+\sum_{j=1}^k\sum_{t=1}^{N_z}\log\tau_{i_j, t}
\end{split}
\end{equation}
where $I_{\{i_1,...,i_k\}}$ is the whole reconstruction given by $\{z_{i_j}^*\}_{j=1}^k$ and $\{\textbf{det}_{i_j}\}_{j=1}^k$, $|I|$ is the cardinality of image $I$. Dropping some constants in Equation \ref{equ:logp_formula}, our loss function is defined as
\begin{equation} 
\label{equ:loss}
L = \|I-I_{\{i_1,...,i_k\}}\|_{vec,2}^2 + \lambda k +\sigma^2 \sum_{j=1}^k[\|\mu_{i_j}\|_2^2+\displaystyle \sum_{t=1}^{N_z} \tau_{i_j, t}^2-2 \sum_{t=1}^{N_z}\log\tau_{i_j, t}],
\end{equation}
where $\lambda = 2\sigma^2 \lambda_0$ can be regarded as a penalty on the number of selected boxes $k$.

If there are $N$ detections in total, it is impossible to enumerate and evaluate all possible ordered subsets $\{i_1,...,i_k\}$.  So we propose a greedy search, see Algorithm \ref{alg:greedy}, to find a good subset in polynomial time.

In Algorithm \ref{alg:wr} we processed selected detections by their occlusion scores because we want to find the visible pixels for each bounding box. 
However, some very low quality detections may have high occlusion score.
Because higher objectness scores indicate more confident detections, and more confident detections are more likely to appear in our final selection, in Algorithm \ref{alg:greedy} we process the detections $\{\textbf{det}_{i_j}\}_{j=1}^k$ according to their objectness scores from high to low. At each step,
given a subset of detections, we pass it to Algorithm \ref{alg:wr} where the detections are reordered according to occlusion score to provide the whole reconstruction and provide the NLL.

We use $S$ to represent the currently selected detections, which is  $\emptyset$ in the beginning. If selecting $S \cup \{\textbf{det}_i\}$ yields smaller loss than with $S$, we prefer interpretation $S \cup \{\textbf{det}_i\}$ to $S$. But we also consider the case when there is a $\textbf{det}_j$, $j<i$, which has significant overlap with $\textbf{det}_i$. It is possible that $\textbf{det}_i$ is the correct detection and $\textbf{det}_j$ isn't. So we select the detection $j< i$ with highest overlap with $i$, and consider the interpretation $(S \setminus \{\textbf{det}_j\})\cup \{\textbf{det}_i\}$. Thus, we compare $S$, $S \cup \{\textbf{det}_i\}$ and $(S \setminus \{\textbf{det}_j\})\cup \{\textbf{det}_i\}$, the one which has the smallest loss is used as the new $S$. Then we move on to the next detection in the objectness score ordering.

\begin{algorithm}
\caption{Detection Selection Algorithm - Greedy Search}\label{alg:greedy}
\KwIn{Image $I$, detection results $\{\textbf{det}_i=(score_i, bb_i, occ_i, cls_i)\}_{i=1}^N$ sorted by $score_i$ from high to low, the penalty $\lambda \geq 0$, assumed variance $\sigma^2>0$, latent dimension $N_z$}
$S_0 \gets \emptyset$\;
$ReconDict \gets \{\}$\;
$L_0 \gets \infty$\;
\For{$i=1$ \KwTo $N$}{
  $(I_{i,1}, ReconDict) \gets WholeReconstruction(S_{i-1}\cup \{\textbf{det}_i\}, ReconDict)$\;
  $L_{i,1} \gets \|I-I_{i,1}\|_2^2+\lambda |S_{i-1}\cup \{\textbf{det}_i\}|+ \displaystyle \sum_{i_j:\textbf{det}_{i_j}\in S_{i-1}\cup \{\textbf{det}_i\} }\sigma^2[\|\mu_{i_j}\|_2^2+\displaystyle \sum_{t=1}^{N_z} \tau_{i_j, t}^2-2 \sum_{t=1}^{N_z}\log\tau_{i_j, t}]$
  \Comment*[r]{$|\cdot|$ is the cardinality of the set}

  \eIf{$S_{i-1}=\emptyset$ or $\max\limits_{\textbf{det}_{j_1} \in S_{i-1}}IoU(\textbf{det}_{j_1},\textbf{det}_i)=0$}{
    $L_{i,2} \gets \infty$\;
  }{
  $\textbf{det}_j = \argmax\limits_{\textbf{det}_{j_1} \in S_{i-1}}IoU(\textbf{det}_{j_1},\textbf{det}_i)$\;
  $(I_{i,2}, ReconDict) \gets WholeReconstruction((S_{i-1}\setminus \{\textbf{det}_j\} )\cup \{\textbf{det}_i\}, ReconDict)$\;
  $L_{i,2} \gets  \|I-I_{i,2}\|_2^2+\lambda |(S_{i-1}\setminus \{\textbf{det}_j\} )\cup \{\textbf{det}_i\}|+ \displaystyle \sum_{i_j:\textbf{det}_{i_j}\in (S_{i-1}\setminus \{\textbf{det}_j\} )\cup \{\textbf{det}_i\} }\sigma^2[\|\mu_{i_j}\|_2^2+\displaystyle \sum_{t=1}^{N_z} \tau_{i_j, t}^2-2 \sum_{t=1}^{N_z}\log\tau_{i_j, t}]$\; 
  }

  $L_i \gets \min(L_{i-1}, L_{i,1}, L_{i,2})$\;
  \uIf{$L_i=L_{i-1}$}{
    $S_i \gets S_{i-1}$\;
  }\uElseIf{$L_i=L_{i,1}$}{
    $S_i \gets S_{i-1}\cup \{\textbf{det}_i\}$\;
  }\Else{$S_i \gets (S_{i-1}\setminus \{\textbf{det}_j\} )\cup \{\textbf{det}_i\}$\;}
}

\KwOut{Selected detections $S_N$}
\end{algorithm}

In Algorithm \ref{alg:greedy}, $\textbf{det}_j$ is chosen as the previously selected detection which has the highest IoU with $\textbf{det}_i$. If there is no previous selected detection, or if all previously selected detections have zero IoU with $\textbf{det}_i$, then $\textbf{det}_j$ doesn't exist and we don't need to consider $(S \setminus \{\textbf{det}_j\})\cup \{\textbf{det}_i\}$. In this case we set $L_{i,2} = \infty$ for that $i$ in Algorithm \ref{alg:greedy}.

Our Detections Selection Algorithm (DSA) greedily selects detection subsets to minimize Equation \ref{equ:loss}. The first term in Equation \ref{equ:loss} encourages DSA to select the interpretation which has better reconstruction performance. At times, selecting duplicated detections can give us almost the same reconstruction loss. The penalty $\lambda$ on the number of boxes is needed to help DSA avoid such situations.

After all the detections have been processed, we choose the final selected detections $S_N$ in Algorithm \ref{alg:greedy} as our interpretation for the image. 

\begin{figure}[t]
  \centering
  \includegraphics[width=\linewidth]{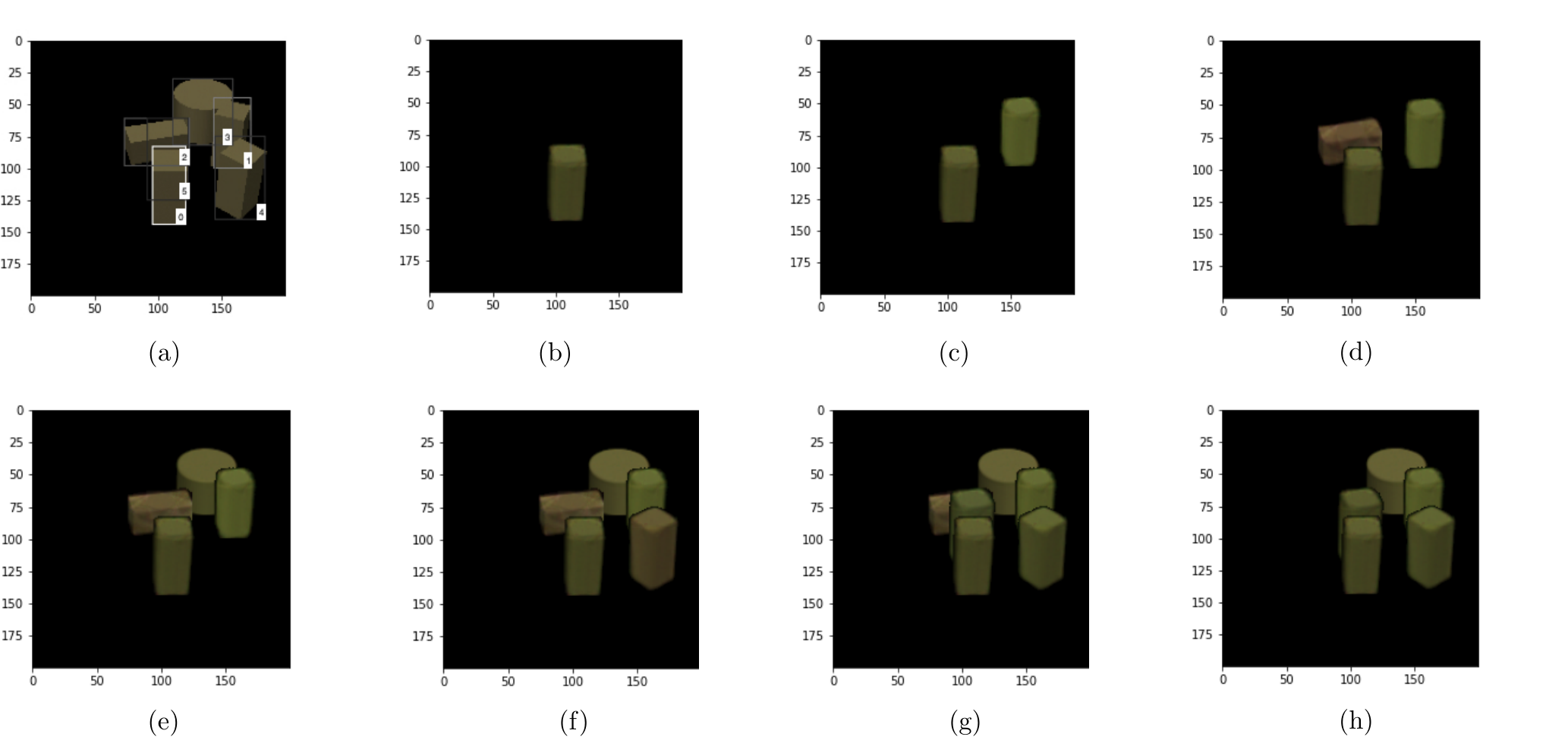}
  \caption{Example of Detection Selection Algorithm.}
  \label{fig:dsa3}
\end{figure}

In Figure \ref{fig:dsa3}, $(a)$ we show the original image with five actual objects and $6$ detections indexed from $0$ to $5$, ordered according to their objectness score, same as in Figure \ref{fig:wr1}. $N=6$ and the last bounding box is redundant. As stated in Algorithm \ref{alg:greedy}, we start from $S_0 = \emptyset$. In the first step, we only consider detection $\{0\}$, which has highest objectness score. The loss is $1307.8$, so we have $S_1=\{0\}$, as shown in $(b)$. Next we move on to bounding box $1$ as the next highest objectness score. Because no bounding box has an intersection with bounding box $1$, in the second step we only consider the ordered set $\{0,1\}$, (there is no possible detection to omit). The loss is $982.5$, which is better than $1307.8$, so $S_2=\{0,1\}$. Its canvas is shown in $(c)$. In the third step bounding box $0$ has the largest IoU with bounding box $2$, so we compare both $\{0,1,2\}$ and $\{1,2\}$ and select $S_3=\{0,1,2\}$ in $(d)$. Its loss is $713.9$. Similarly we compare $\{0,1,2,3\}$ and $\{0,2,3\}$ and select $S_4=\{0,1,2,3\}$ in $(e)$ which gives us loss $356.3$. In step $5$, $\{0,1,2,3,4\}$ and $\{0,2,3,4\}$ are compared and $S_5=\{0,1,2,3,4\}$ with loss $151.6$ is selected. Finally, since bounding box $2$ has the largest IoU with bounding box $5$, we process $\{0,1,2,3,4,5\}$ and $\{0,1,3,4,5\}$, which gives us losses $165.2$ and $253.0$ respectively, and their canvases are shown in $(g), (h)$. But neither $\{0,1,2,3,4,5\}$ or $\{0,1,3,4,5\}$ has a smaller loss than $S_5$, so $S_6=S_5=\{0,1,2,3,4\}$. Therefore, ultimately we interpret the image by bounding boxes $\{0,1,2,3,4\}$, which means there are $5$ objects and our predicted labels are the corresponding labels of the bounding boxes $\{0,1,2,3,4\}$. Some more experiments about DSA are shown in Section \ref{exp:dsa}.

\section{Data}
\label{sec:data}
\begin{figure}[t]
  \centering
  \includegraphics[width=\linewidth]{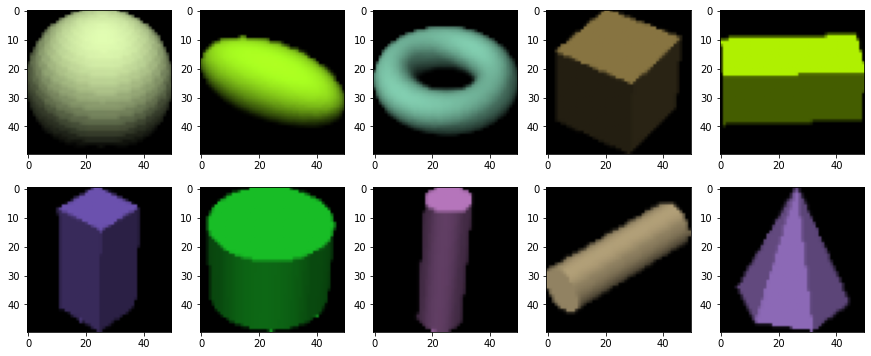}
  \caption{10 classes of objects.}
  \label{fig:10classes}
\end{figure}

Our synthesized datasets have $10$ classes of objects labeled from class $1$ through class $10$. The $10$ classes are shown in Figure \ref{fig:10classes}: sphere, ellipsoid, torus, regular cube, lying thin cuboid, standing thin cuboid, regular cylinder, standing thin cylinder, lying thin cylinder and cone. Regardless of the object classes, each image has a random color $(r,g,b)$ on all of its objects, where $0<=r,g,b<=1$ are uniformly chosen conditioned on $r+g+b>=1$. We set the objects to have the same color to make the detection more difficult.

In all our datasets, the objects are placed onto a black background by randomly choosing their locations. We use the Python package 'pyvista' to generate the images. In each image, there are three lights with fixed directions and fixed intensities as $0.5, 0.5, 0.2$. The camera position and focus are fixed as well, but the objects can rotate from $0$ to $360$ degrees horizontally. It can be seen from Figure \ref{fig:10classes} that classes $1, 3, 7, 8$ are rotational invariant, others are not.

\subsection{Training sets}

In order to train the Faster R-CNN model or  Faster R-CNN-OC model in section \ref{sec:ob}, we generate paired occluded objects on a $200\times 200$ black background as our training set. These objects are  placed on an invisible floor. Rejection sampling is applied to ensure that although they are occluded, the two objects won't be too close to each other and each object has at least $200$ visible pixels. The classes of the two paired objects are chosen from the $10$ classes we have to make each class appears exactly $1000$ times in the dataset. So we have $1000\times 10/2=5000$ images in total for Faster R-CNN.

\begin{figure}[t]
  \centering
  \includegraphics[width=\linewidth]{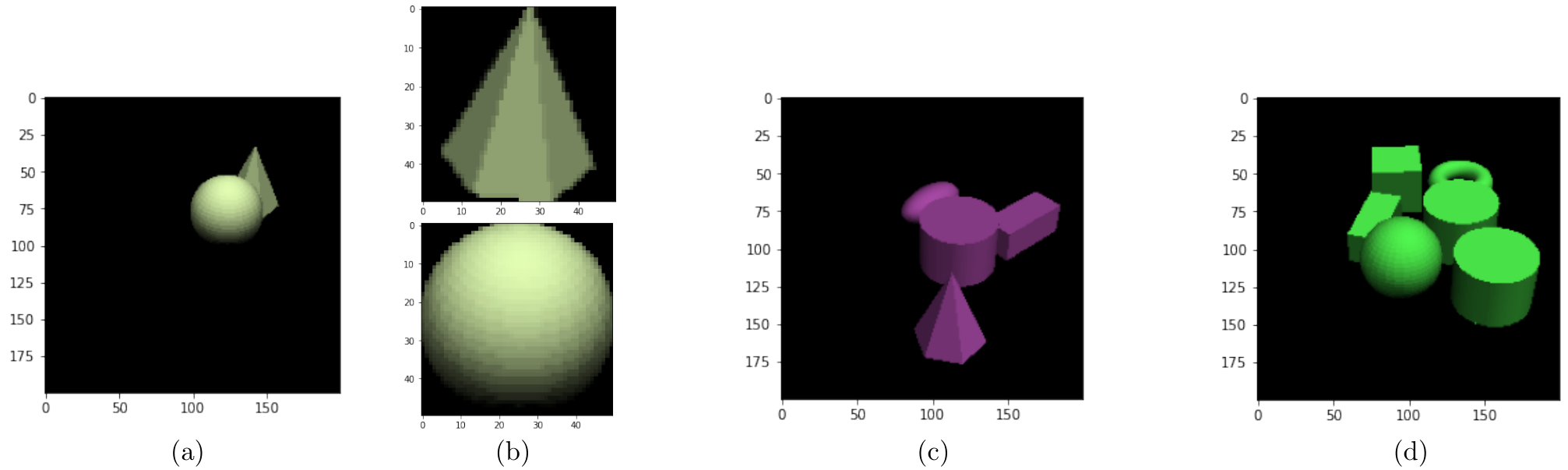}
  \caption{Datasets.}
  \label{fig:datasets}
\end{figure}

For the single reconstruction algorithm, we train a VAE decoder. When we generate the pairs of objects for the Faster R-CNN training images, we keep the same individual objects as our decoder training data, but they are isolated and centered in a $50\times 50$ images. Another difference is that, the isolated objects are re-scaled to make them as large as possible in the $50\times 50$ images. This is implemented using the parameterized sampling grid technique mentioned in section \ref{subsubsec:arb_size_recon}. In Figure \ref{fig:datasets}, image $(a)$ is a training image for Faster R-CNN and two images in $(b)$ are the corresponding training data for the decoder. We have $5000 \times 2=10000$ images for decoder.

\subsection{Validation and test sets}

To test the performance of our post processing methods, we have a validation set and a test set. These datasets still contain $200\times 200$ images, but in each image we have $3$ to $7$ objects. The difficulty of post processing usually increases as the number of objects increases. 

Both our validation set and test set for post processing have $500$ images. We ensure that every object has at least $200$ visible pixels and objects are not too close to each other. For the validation set, we have $250$ images with $3$ objects and $250$ images with $4$ objects in them. An example of a validation image can be found in image $(c)$ of Figure \ref{fig:datasets}. For the test set, we have $150, 150, 200$ images with $5, 6, 7$ objects respectively. Image $(d)$ in Figure \ref{fig:datasets} is a test set example which contains $6$ objects.

\section{Experiments}
\label{sec:exp}
\subsection{Implementation Details}
\label{exp:imple}

As explained in section \ref{sec:data}, we train the Faster R-CNN and our Faster R-CNN-OC model with $5000$ paired occluded objects. We further divide them into $4000$ training images and $1000$ validation images. Both models are trained with $batch\_size=10$ and the default Faster R-CNN parameters without any pre-training. The training stops when we observe $10$ continuous epochs of no improvement on the validation loss. The Faster R-CNN and Faster R-CNN-OC model stopped at epoch $79$ and $64$ respectively. For the occlusion branch, during training the "occlusion scores" of the upper object and the lower object are set to be $1.0$ and $0.0$ respectively. 

We train a separate decoder for each class. The latent dimension is $10$. There is one hidden layer with $300$ hidden units fully connected to a layer with 7500 units corresponding to the 50x50x3 output. We use a ReLU nonlinearity after the hidden layer and a sigmoid nonlinearity after the final layer. The first $8000$ images are used in training the decoder and the remaining $2000$ images are used for testing. Our  decoder is trained for $400$ epochs and we update the decoder parameters once after $10$ optimization iterations of the latent code. We fix $\sigma=0.1$ and $batch\_size=100$. Adam optimizers are used and the learning rates for updating decoder parameters and latent code are $0.0001$ and $0.01$.

For the full detection selection algorithm (DSA), we drop detections with objectness score less than $0.25$, such detections would anyhow be rejected by the greedy search but at an unnecessary computational cost.

\subsection{Experiments on Occlusion Scores}
\label{exp:occ}

\begin{figure}[t]
  \centering
  \includegraphics[width=\linewidth]{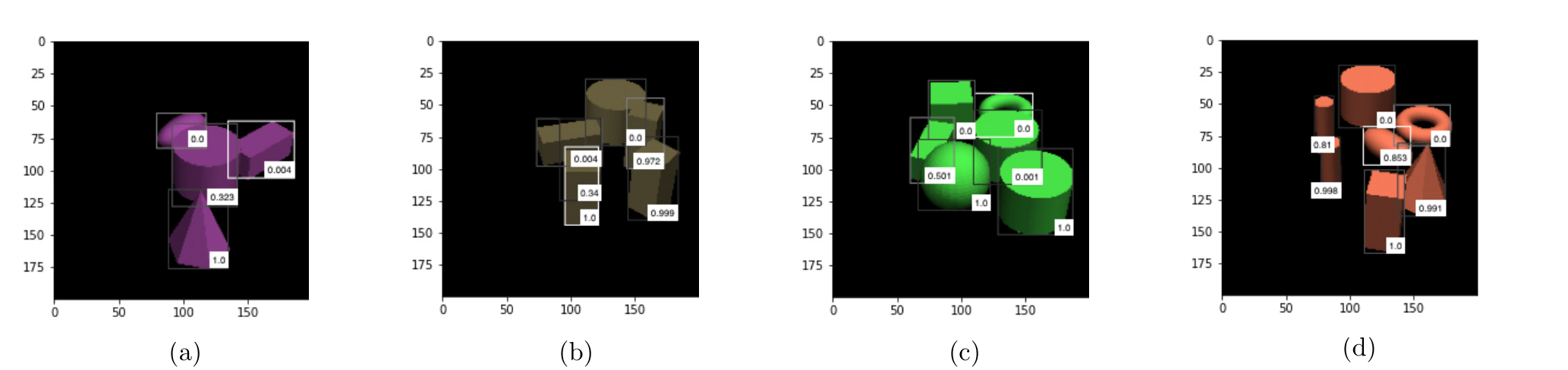}
  \caption{Example of Predicted Occlusion Scores.}
  \label{fig:occlusion_score}
\end{figure}

The Faster R-CNN-OC model is trained on pairs of objects, but it generalizes well when we test on three or more overlapping objects. Since occlusion relationship reasoning is not our main focus, we only display some results in Figure \ref{fig:occlusion_score}, where the predicted occlusion score is displayed at the lower right corner for each bounding box. Note that for clarity we only display the top several bounding boxes after NMS by Faster R-CNN. 

\subsection{Experiments on DSA}
\label{exp:dsa}
People often use mAP (mean Average Precision) to evaluate the quality of detections. However, in order to determine the correct number of objects, the precision when $recall=1$ matters the most. In this work we use two types of accuracies as our evaluation metrics:
\begin{itemize}
    \item The precent of images where the correct number of boxes is chosen. 
    
    \item The percent of images with the correct number of boxes and  correct predicted labels.
\end{itemize}

For example, if there are one class-1 object and two class-2 objects in the image, and our prediction is two class-1 objects plus one class-2 object, then we are correct under the first evaluation metric and wrong under the second evaluation metric. Clearly, the second evaluation metric is more strict.

\subsubsection{NMS, Soft-NMS and DIoU-NMS}

In Table \ref{tab:tab1} we compare three post processing methods for the Faster R-CNN: NMS, Soft-NMS and DIoU-NMS. The parameters $T_{boxes}$ and $T_{labels}$ are both thresholds mentioned in section \ref{sec:intro}. In the first half of the table, we determined the optimal threshold $T_{boxes}$ using a grid search over the range $0.01, 0.02, ..., 0.99$ as threshold on the validation set. Detections above the chosen threshold for each method are used as final detections. Similarly, $T_{labels}$ tries to maximize the accuracy of labels in the validation set. It seems that the chosen $T_{boxes}$ and $T_{labels}$ are pretty close to each other. The thresholds for Soft-NMS are lower because Soft-NMS decreases  objectness scores. For comparison, in the second half of Table \ref{tab:tab1}, we just fix the thresholds to be $0.5$. Due to these lower thresholds, the accuracies of NMS decrease drastically.

\begin{table}[th]
  \caption{NMS, Soft-NMS and DIoU-NMS with Faster R-CNN}
  \label{tab:tab1}
  \centering
  \begin{tabular}{lcccc}
    \toprule
    \multicolumn{1}{c}{\textbf{Methods}} &
    
    \multicolumn{1}{c}{\textbf{$T_{boxes}$}} & 
        \multicolumn{1}{c}{\textbf{$T_{labels}$}}&
                                         
        \multicolumn{1}{c}{\textbf{Accuracy for Boxes}}&
     \multicolumn{1}{c}{\textbf{Accuracy for Labels}}\\
    \midrule
NMS  & 0.91 & 0.91 & 0.962 (0.0086) & 0.962 (0.0086) \\
Soft-NMS & 0.69 & 0.69 & 0.950 (0.0097)  & 0.950 (0.0097) \\
DIoU-NMS  & 0.91 & 0.91 & 0.940 (0.0106) & 0.940 (0.0106) \\
\midrule

NMS  & 0.5 & 0.5 & 0.772 (0.0188) & 0.772 (0.0188) \\
Soft-NMS & 0.5 & 0.5 & 0.946 (0.0101) & 0.946 (0.0101) \\
DIoU-NMS  & 0.5 & 0.5 & 0.934 (0.0111) & 0.934 (0.0111) \\
    \bottomrule
  \end{tabular}
  
\end{table}

The accuracies in Table \ref{tab:tab1} are calculated on the test set based on the thresholds obtained by the validation set. Because the validation set and test set have different distributions, the thresholds may not be optimal. The numbers in the parenthesis are the estimated standard deviations using $\sqrt{\frac{\hat{p}(1-\hat{p})}{n}}$, where $\hat{p}$ is the average accuracy and $n=500$ is the number of test samples. Accuracy for boxes and accuracy for labels are the first and second evaluation metrics mentioned earlier. We can see that for every method, the accuracy for boxes and labels are the same.

\subsubsection{NMS+DSA and Soft-NMS+DSA using Faster R-CNN-OC}
Given a set of detections by an object detection algorithm,
DSA as described has no ability of reducing False Negatives, but can be used to reduce False Positives. We use DSA after NMS or DSA after Soft-NMS, to post-process the detections of the Faster R-CNN-OC. The threshold for NMS is set at .5. In Table \ref{tab:tab2}, DSA after NMS is called "NMS+DSA" and DSA after Soft-NMS is called "Soft-NMS+DSA".

The penalties $\lambda_{boxes}$ and $\lambda_{labels}$ are also chosen by the validation set to maximize the two evaluation metrics respectively. We tried  the values $10, 20, 30, 40, 50$. If there are ties, we choose the median. 

\begin{table}[th]
  \caption{NMS+DSA and Soft-NMS+DSA on Faster R-CNN-OC}
  \label{tab:tab2}
  \centering
  \begin{tabular}{lcccc}
    \toprule
    \multicolumn{1}{c}{\textbf{Methods}} &
    
    \multicolumn{1}{c}{\textbf{$\lambda_{boxes}$}} & 
    \multicolumn{1}{c}{\textbf{$\lambda_{labels}$}} & 
                                         
        \multicolumn{1}{c}{\textbf{Accuracy for Boxes}}&
     \multicolumn{1}{c}{\textbf{Accuracy for Labels}}\\
    \midrule
  NMS+DSA   & 15 & 15 & 0.980 (0.0063) &  0.980 (0.0063) \\
  Soft-NMS+DSA  & 20 & 20 & 0.982 (0.0059) & 0.980 (0.0063) \\
    \bottomrule
  \end{tabular}
  
\end{table}

In Table \ref{tab:tab2}, we observe clear improvements over the original NMS or Soft-NMS in Table \ref{tab:tab1}. NMS+DSA performs especially well.
\subsubsection{Recovering False Negatives}

\begin{figure}[t]
  \centering
  \includegraphics[width=1.6in]{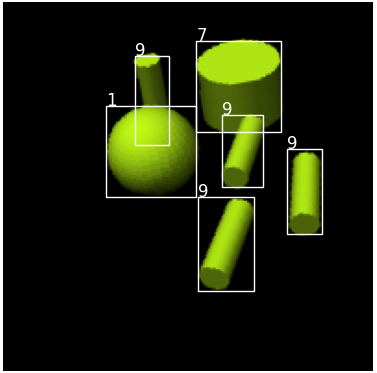}
  \includegraphics[width=1.61in]{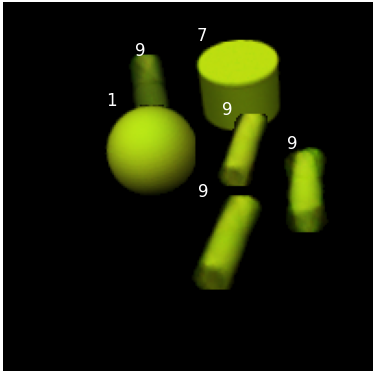}
  \includegraphics[width=1.6in]{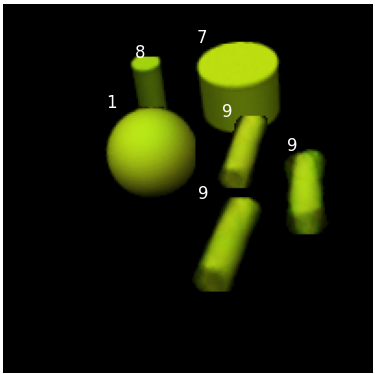}
  \caption{Left: Faster R-CNN output on rotated image, Middle: Whole reconstruction without class 8 competition, Right: Whole reconstruction with class 8 competition.}
  \label{fig:cl8}
\end{figure}

We conducted a simple experiment to see how DSA might be extended to recover missed detections of NMS or soft-NMS.
We rotated each test image by 10 degrees. This minor perturbation significantly reduces the accuracy of the detection algorithms. For example soft-NMS yields 0.90 for number of boxes and 0.652 for proportion of images with correct labels. Just observing the results it is clear that the small rotation leads the faster R-CNN to label many instances of class 8 - the upright cylinder as class 9. So we added a minor hack in the code, where any time a box is labeled 9, we also run the whole reconstruction algorithm on exactly the same input except that the new box is labeled 8 instead of 9 and then compare the NLL's. Furthermore, we introduce a new variable $\alpha$ in the decoder for rotation  in addition to the translation variable so that the decoder optimization is over $\mu,\Gamma,(t_x,t_y), \alpha$. This yielded a significant improvement of 0.906 proportion of images with correct number of detected boxes and 0.856 proportion of images with all labels correct. In Figure \ref{fig:cl8} we show the different whole reconstructions produced without and with the likelihood comparison between class 9 and class 8.
This experiment points to the possibility of recovering from distribution shifts by extending the free parameters of the decoder as well as entertaining more than one class label for each detected box.

\subsubsection{Enlarged Objects}

Another experiment of interest is rescaling all objects by the same proportion $\frac{10}{9}$ in the test set, and keeping the training and validation sets the same. This is implemented by cropping a $180\times 180$ region which contains all the objects and enlarging it to become a $200\times 200$ image.

\begin{table}[th]
  \caption{NMS, Soft-NMS and DIoU-NMS with Faster R-CNN on Enlarged Objects}
  \label{tab:tab3}
  \centering
  \begin{tabular}{lcccc}
    \toprule
    \multicolumn{1}{c}{\textbf{Methods}} &
    
    \multicolumn{1}{c}{\textbf{$T_{boxes}$}} & 
        \multicolumn{1}{c}{\textbf{$T_{labels}$}}&
                                         
        \multicolumn{1}{c}{\textbf{Accuracy for Boxes}}&
     \multicolumn{1}{c}{\textbf{Accuracy for Labels}}\\
    \midrule
NMS  & 0.91 & 0.91 & 0.886 (0.0142)  & 0.880 (0.0145) \\
Soft-NMS & 0.69 & 0.69 & 0.916 (0.0124)  & 0.908 (0.0129) \\
DIoU-NMS  & 0.91 & 0.91 & 0.886 (0.0142) & 0.874 (0.0148) \\
    \bottomrule
  \end{tabular}
  
\end{table}

\begin{table}[th]
  \caption{NMS+DSA and Soft-NMS+DSA on Enlarged Objects with Faster R-CNN-OC}
  \label{tab:tab4}
  \centering
  \begin{tabular}{lcccc}
    \toprule
    \multicolumn{1}{c}{\textbf{Methods}} &
    
    \multicolumn{1}{c}{\textbf{$\lambda_{boxes}$}} & 
    \multicolumn{1}{c}{\textbf{$\lambda_{labels}$}} & 
                                         
        \multicolumn{1}{c}{\textbf{Accuracy for Boxes}}&
     \multicolumn{1}{c}{\textbf{Accuracy for Labels}}\\
    \midrule
  NMS+DSA   & 15 & 15 & 0.988 (0.0049) & 0.982 (0.0059)  \\
  Soft-NMS+DSA  & 20 & 20 & 0.964 (0.0083) & 0.960 (0.0088) \\
    \bottomrule
  \end{tabular}
  
\end{table}

The results are summarized in Table \ref{tab:tab3} and Table \ref{tab:tab4}. From the results we can see that DSA leads to highly significant improvements. NMS+DSA again performs especially well. From ordinary objects to the enlarged objects, the accuracies of NMS drop from $0.962, 0.962$ to $0.886, 0.880$, while the accuracies of NMS+DSA don't drop. 

Because the sizes of objects in the new images are new to Faster R-CNN, the objectness scores become less reliable. That's why NMS, Soft-NMS and DIoU-NMS exhibit worse performance. Our single reconstruction algorithm is capable of handling bounding boxes of various scales, so our DSA still works on enlarged objects. 

\begin{figure}[t]
  \centering
  \includegraphics[width=1.6in]{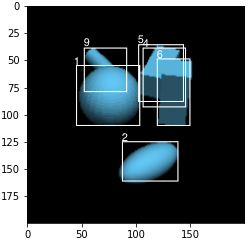}
  \includegraphics[width=1.61in]{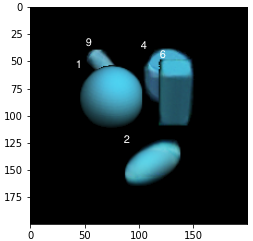}
  \includegraphics[width=1.6in]{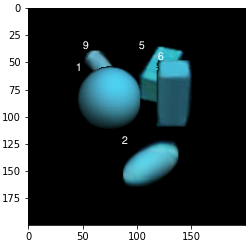}
  \caption{Left: Faster R-CNN output , Middle: Whole reconstruction of top 5 bounding boxes, Right: Whole reconstruction of top 4 and the 6th bounding boxes.}
  \label{fig:backtrack}
\end{figure}

Figure \ref{fig:backtrack} illustrates why we need to compare $(S \setminus \{\textbf{det}_j\})\cup \{\textbf{det}_i\}$ with $S$ and $S \cup \{\textbf{det}_i\}$ in the detection selection algorithm. In Figure \ref{fig:backtrack}, an object of class 5 is predicted in two different bounding boxes by Faster R-CNN as class 4 and class 5 with objectness scores $0.89$ and $0.82$ respectively. Thus, the class 4 object is processed before the class 5 object in the DSA. The whole reconstruction by the top 5 detections in terms of objectness scores yields loss $430.36$. It selects the wrong bounding box of class 4. In the next step, DSA considers dropping the bounding box of class 4 and adding the bounding box of class 5. The loss decreases to $376.92$, and it gives us the right interpretation.

\section{Discussion}
\label{sec:conc}
In this work, we have proposed the Detection Selection Algorithm (DSA) and several supportive algorithms, which can be used to determine the exact number of objects and their labels in the image. DSA is used after NMS or related methods. The framework is likelihood based where comparisons are made between image interpretations - ordered sequences of instantiated objects. The probabilistic framework offers a global evaluation of any interpretation and takes into account the relationships between the different objects. Some byproducts of DSA include finding the occlusion sequence of objects, reconstructing the invisible parts of objects, and generating images on a given set of hypothesized objects.

We note that most network models used today in image processing are fully feed-forward. The input passes through the network and produces the output. This works well when there is ample training data and when the distribution of the test data is the same as that of the training data, i.e. no distributional shift. However such methods are quite sensitive to distributional shifts as demonstrated in the experiments above, and it appears to us that in certain settings adjusting to such shifts without retraining necessitates an online optimization procedure that can accommodate the modified distribution, and in particular using global likelihood based reasoning. A full probabilistic model is the most principled way to achieve this, albeit at a significant computational cost. Our greedy  algorithm implements only one step back search, only inspecting the detection with highest overlap. More extensive searches could be implemented exploring a wider range of ordered subsets of the detections, again, at a higher computational cost.

Further research is needed to extend the DSA to real-world images with colored background and clutter. A possible way to do this is to extract features from real-world images in such a way that reduce the task to dealing with images on black background.

\bibliographystyle{IEEEtran}
\bibliography{mybib}

% Generated by IEEEtran.bst, version: 1.14 (2015/08/26)
\begin{thebibliography}{10}
\providecommand{\url}[1]{#1}
\csname url@samestyle\endcsname
\providecommand{\newblock}{\relax}
\providecommand{\bibinfo}[2]{#2}
\providecommand{\BIBentrySTDinterwordspacing}{\spaceskip=0pt\relax}
\providecommand{\BIBentryALTinterwordstretchfactor}{4}
\providecommand{\BIBentryALTinterwordspacing}{\spaceskip=\fontdimen2\font plus
\BIBentryALTinterwordstretchfactor\fontdimen3\font minus
  \fontdimen4\font\relax}
\providecommand{\BIBforeignlanguage}[2]{{%
\expandafter\ifx\csname l@#1\endcsname\relax
\typeout{** WARNING: IEEEtran.bst: No hyphenation pattern has been}%
\typeout{** loaded for the language `#1'. Using the pattern for}%
\typeout{** the default language instead.}%
\else
\language=\csname l@#1\endcsname
\fi
#2}}
\providecommand{\BIBdecl}{\relax}
\BIBdecl

\bibitem{faster}
S.~Ren, K.~He, R.~Girshick, and J.~Sun, ``Faster r-cnn: Towards real-time
  object detection with region proposal networks,'' in \emph{Proceedings of the
  28th International Conference on Neural Information Processing Systems -
  Volume 1}, ser. NIPS'15.\hskip 1em plus 0.5em minus 0.4em\relax Cambridge,
  MA, USA: MIT Press, 2015, p. 91–99.

\bibitem{softNMS}
N.~Bodla, B.~Singh, R.~Chellappa, and L.~S. Davis, ``Soft-nms--improving object
  detection with one line of code,'' in \emph{Proceedings of the IEEE
  international conference on computer vision}, 2017, pp. 5561--5569.

\bibitem{amit2007pop}
Y.~Amit and A.~Trouv{\'e}, ``Pop: Patchwork of parts models for object
  recognition,'' \emph{International Journal of Computer Vision}, vol.~75,
  no.~2, pp. 267--282, 2007.

\bibitem{FastRCNN}
\BIBentryALTinterwordspacing
R.~B. Girshick, ``Fast {R-CNN},'' \emph{CoRR}, vol. abs/1504.08083, 2015.
  [Online]. Available: \url{http://arxiv.org/abs/1504.08083}
\BIBentrySTDinterwordspacing

\bibitem{SSD}
\BIBentryALTinterwordspacing
W.~Liu, D.~Anguelov, D.~Erhan, C.~Szegedy, S.~E. Reed, C.~Fu, and A.~C. Berg,
  ``{SSD:} single shot multibox detector,'' \emph{CoRR}, vol. abs/1512.02325,
  2015. [Online]. Available: \url{http://arxiv.org/abs/1512.02325}
\BIBentrySTDinterwordspacing

\bibitem{YOLO}
\BIBentryALTinterwordspacing
J.~Redmon, S.~K. Divvala, R.~B. Girshick, and A.~Farhadi, ``You only look once:
  Unified, real-time object detection,'' \emph{CoRR}, vol. abs/1506.02640,
  2015. [Online]. Available: \url{http://arxiv.org/abs/1506.02640}
\BIBentrySTDinterwordspacing

\bibitem{ENMS}
A.~Neubeck and L.~Van~Gool, ``Efficient non-maximum suppression,'' in
  \emph{18th International Conference on Pattern Recognition (ICPR'06)},
  vol.~3.\hskip 1em plus 0.5em minus 0.4em\relax IEEE, 2006, pp. 850--855.

\bibitem{review_NMS}
M.~Gong, D.~Wang, X.~Zhao, H.~Guo, D.~Luo, and M.~Song, ``A review of
  non-maximum suppression algorithms for deep learning target detection,'' in
  \emph{Seventh Symposium on Novel Photoelectronic Detection Technology and
  Applications}, vol. 11763.\hskip 1em plus 0.5em minus 0.4em\relax SPIE, 2021,
  pp. 821--828.

\bibitem{softerNMS}
Y.~He, X.~Zhang, M.~Savvides, and K.~Kitani, ``Softer-nms: Rethinking bounding
  box regression for accurate object detection,'' \emph{arXiv preprint
  arXiv:1809.08545}, vol.~2, no.~3, pp. 69--80, 2018.

\bibitem{IoU-GuidedNMS}
B.~Jiang, R.~Luo, J.~Mao, T.~Xiao, and Y.~Jiang, ``Acquisition of localization
  confidence for accurate object detection,'' in \emph{Proceedings of the
  European conference on computer vision (ECCV)}, 2018, pp. 784--799.

\bibitem{AdaptiveNMS}
S.~Liu, D.~Huang, and Y.~Wang, ``Adaptive nms: Refining pedestrian detection in
  a crowd,'' in \emph{Proceedings of the IEEE/CVF conference on computer vision
  and pattern recognition}, 2019, pp. 6459--6468.

\bibitem{DIOU-NMS}
Z.~Zheng, P.~Wang, W.~Liu, J.~Li, R.~Ye, and D.~Ren, ``Distance-iou loss:
  Faster and better learning for bounding box regression,'' in
  \emph{Proceedings of the AAAI conference on artificial intelligence},
  vol.~34, no.~07, 2020, pp. 12\,993--13\,000.

\bibitem{NMS-face}
E.~Zaytseva and J.~Vitri{\`a}, ``A search based approach to non maximum
  suppression in face detection,'' in \emph{2012 19th IEEE International
  Conference on Image Processing}.\hskip 1em plus 0.5em minus 0.4em\relax IEEE,
  2012, pp. 1469--1472.

\bibitem{HS-NMS}
Y.~Song, Q.-K. Pan, L.~Gao, and B.~Zhang, ``Improved non-maximum suppression
  for object detection using harmony search algorithm,'' \emph{Applied Soft
  Computing}, vol.~81, p. 105478, 2019.

\bibitem{probabilisticFRCNN}
D.~Yi, J.~Su, and W.-H. Chen, ``Probabilistic faster r-cnn with stochastic
  region proposing: Towards object detection and recognition in remote sensing
  imagery,'' \emph{Neurocomputing}, vol. 459, pp. 290--301, 2021.

\bibitem{MT-ORL}
P.~Feng, Q.~She, L.~Zhu, J.~Li, L.~Zhang, Z.~Feng, C.~Wang, C.~Li, X.~Kang, and
  A.~Ming, ``Mt-orl: Multi-task occlusion relationship learning,'' in
  \emph{Proceedings of the IEEE/CVF International Conference on Computer
  Vision}, 2021, pp. 9364--9373.

\bibitem{ORM}
X.~Yuan, A.~Kortylewski, Y.~Sun, and A.~Yuille, ``Robust instance segmentation
  through reasoning about multi-object occlusion,'' in \emph{Proceedings of the
  IEEE/CVF Conference on Computer Vision and Pattern Recognition}, 2021, pp.
  11\,141--11\,150.

\bibitem{seGAN}
K.~Ehsani, R.~Mottaghi, and A.~Farhadi, ``Segan: Segmenting and generating the
  invisible,'' in \emph{Proceedings of the IEEE conference on computer vision
  and pattern recognition}, 2018, pp. 6144--6153.

\bibitem{GAN}
\BIBentryALTinterwordspacing
I.~J. Goodfellow, J.~Pouget-Abadie, M.~Mirza, B.~Xu, D.~Warde-Farley, S.~Ozair,
  A.~Courville, and Y.~Bengio, ``Generative adversarial networks,'' 2014.
  [Online]. Available: \url{https://arxiv.org/abs/1406.2661}
\BIBentrySTDinterwordspacing

\bibitem{KINS}
L.~Qi, L.~Jiang, S.~Liu, X.~Shen, and J.~Jia, ``Amodal instance segmentation
  with kins dataset,'' in \emph{Proceedings of the IEEE/CVF Conference on
  Computer Vision and Pattern Recognition}, 2019, pp. 3014--3023.

\bibitem{girshick2015fast}
R.~Girshick, ``Fast r-cnn,'' 2015.

\bibitem{kingma2013auto}
D.~P. Kingma and M.~Welling, ``Auto-encoding variational bayes,'' \emph{arXiv
  preprint arXiv:1312.6114}, 2013.

\bibitem{SPN}
M.~Jaderberg, K.~Simonyan, A.~Zisserman, and K.~Kavukcuoglu, ``Spatial
  transformer networks,'' in \emph{Proceedings of the 28th International
  Conference on Neural Information Processing Systems - Volume 2}, ser.
  NIPS'15.\hskip 1em plus 0.5em minus 0.4em\relax Cambridge, MA, USA: MIT
  Press, 2015, p. 2017–2025.

\bibitem{hoffman2013stochastic}
M.~D. Hoffman, D.~M. Blei, C.~Wang, and J.~Paisley, ``Stochastic variational
  inference,'' \emph{Journal of Machine Learning Research}, 2013.

\bibitem{gregor2015draw}
K.~Gregor, I.~Danihelka, A.~Graves, D.~Rezende, and D.~Wierstra, ``Draw: A
  recurrent neural network for image generation,'' in \emph{International
  conference on machine learning}.\hskip 1em plus 0.5em minus 0.4em\relax PMLR,
  2015, pp. 1462--1471.

\end{thebibliography}

%\appendix

\end{document}